\def\year{2021}\relax
\documentclass[letterpaper]{article} 
\usepackage{aaai21}  
\usepackage{times}  
\usepackage{helvet} 
\usepackage{courier}  
\usepackage[hyphens]{url}  
\usepackage{graphicx} 
\usepackage{natbib}  
\usepackage{caption} 
\usepackage{amsmath}
\usepackage{amssymb}
\usepackage{booktabs}
\usepackage{multirow}

\title{Progressive One-Shot Human Parsing}
\author {

        Haoyu He, \textsuperscript{\rm 1}
        Jing Zhang, \textsuperscript{\rm 1} 
        Bhavani Thuraisingham, \textsuperscript{\rm 2}
        Dacheng Tao \textsuperscript{\rm 1}
        \\
}
\affiliations {
    \textsuperscript{\rm 1} The University of Sydney, Australia,  \\
    \textsuperscript{\rm 2} The University of Texas at Dallas, USA \\
    hahe7688@uni.sydney.edu.au, bxt043000@utdallas.edu,  \{jing.zhang1,dacheng.tao\}@sydney.edu.au
}
\begin{document}
\maketitle

\begin{abstract}
Prior human parsing models are limited to parsing humans into classes pre-defined in the training data, which is not flexible to generalize to unseen classes, $e.g.$, new clothing in fashion analysis. In this paper, we propose a new problem named one-shot human parsing (OSHP) that requires to parse human into an open set of reference classes defined by any single reference example. During training, only base classes defined in the training set are exposed, which can overlap with part of reference classes. In this paper, we devise a novel Progressive One-shot Parsing network (POPNet) to address two critical challenges , $i.e.$, \textbf{\emph{testing bias}} and \textbf{\emph{small sizes}}. POPNet consists of two collaborative metric learning modules named Attention Guidance Module and Nearest Centroid Module, which can learn representative prototypes for base classes and quickly transfer the ability to unseen classes during testing, thereby reducing testing bias. Moreover, POPNet adopts a progressive human parsing framework that can incorporate the learned knowledge of parent classes at the coarse granularity to help recognize the descendant classes at the fine granularity, thereby handling the small sizes issue. Experiments on the ATR-OS benchmark tailored for OSHP demonstrate POPNet outperforms other representative one-shot segmentation models by large margins and establishes a strong baseline. Source code can be found at https://github.com/Charleshhy/One-shot-Human-Parsing.


\end{abstract}

\section{Introduction}
Human parsing is a fundamental visual understanding task, requiring segmenting human images into explicit body parts as well as some clothing classes at the pixel level. It has a broad range of applications especially in the fashion industry including fashion image generating \cite{han2019clothflow}, virtual try-on \cite{dong2019fw}, and fashion image retrieval \cite{wang2017clothing}. Although Convolutional Neural Network (CNN) has made significant progress by leveraging the large-scale human parsing datasets \cite{liang2015human,liang2016semantic2,ruan2019devil}, the parsing results are restricted to the classes pre-defined in the training set, $e.g.$ 18 classes in ATR \cite{liang2015}, and 19 classes in LIP \cite{liang2018look}. However, due to the vast new clothing, fast varying styles in the fashion industry, parsing humans into fixed and pre-defined classes has limited the usage of human parsing models in various downstream applications.

To address the problem, we make the first attempt by defining a new task named One-Shot Human Parsing (OSHP), inspired by one-shot learning \cite{koch2015siamese, vinyals2016matching}. OSHP requires to parse human in a query image into an open set of reference classes defined by any single reference example ($i.e.$, support image), no matter they have been seen during training (base classes) or not (novel classes). In this way, we can flexibly add and remove the novel classes depending on the requirements of specific applications without the need for collecting and annotating new training samples and retraining. 

\begin{figure}[t]
  \centering
  \includegraphics[width=\linewidth]{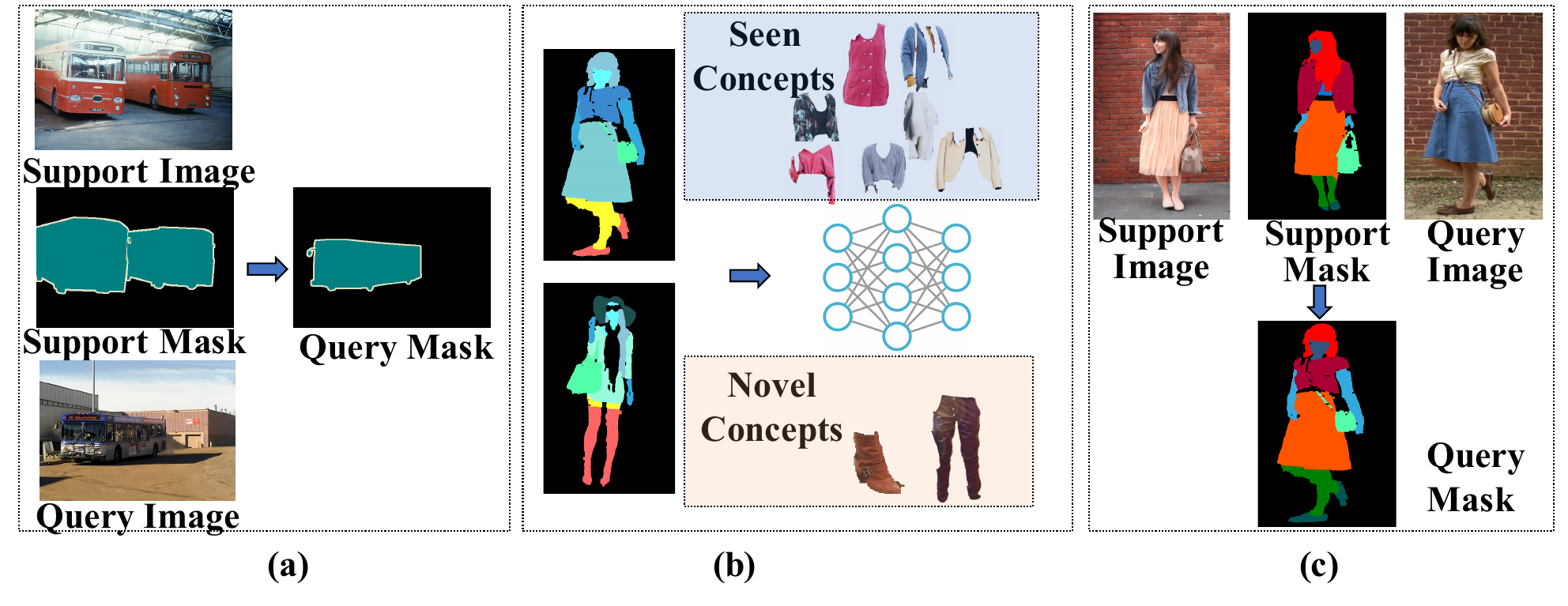}
  \caption{Comparison of OSHP against the OS3 task. (a) The classes in OS3 are large and holistic objects and only novel classes are presented and needed to be recognized during evaluation. (b) In OSHP, both base classes (cold colors) and novel (warm colors) classes are presented and needed to be recognized during the evaluation, leading to the testing bias issue. (c) In OSHP, the part of each class is small and correlated with other parts within the same human foreground.}
  \label{fig:intro}
\end{figure}

One similar task is One-shot semantic segmentation (OS3) \cite{zhang2019canet, zhang2019pyramid, wang2019panet}, that transfers the segmenting knowledge from the pre-defined base classes to the novel classes as shown in Figure~\ref{fig:intro} (a). However, OSHP is more challenging than the OS3 in two ways. Firstly, only novel classes are presented and needed to be recognized during evaluation in OS3, while both base classes and novel classes should be recognized simultaneously during evaluation in OSHP as shown in Figure~\ref{fig:intro} (b), which is indeed a variant of generalized few-shot learning (GFSL) problem \cite{gidaris2018dynamic, ren2019incremental, shi2019relational, ye2019learning}. Note that the two types of classes have imbalanced training data, $i.e.$, there may be many training images for base classes while only a single support image for novel classes. Moreover, since we have no prior information on the explicit definition of novel classes, they are treated as background when presented during the training stage. Consequently, the parsing model may overfit the base classes and specifically lean towards the background for those novel classes, leading to the \textbf{\emph{testing bias}} issue. Secondly, the object in OS3 to be segmented is the intact and salient foreground, while the part of each class that needs to be recognized is small and correlated with other parts within the same human foreground as shown in Figure~\ref{fig:intro}(c), resulting in the \textbf{\emph{small sizes}} issue. Directly deploying OS3 models to OSHP suffers from severe performance degradation due to these two issues.

In this work, we propose a novel POPNet for OSHP. To transfer the learning ability to recognizing base classes in the human body to the novel classes, POPNet employs a dual-metric learning strategy via an Attention Guidance Module (AGM) and Nearest Centroid Module (NCM). AGM aims to learn a discriminative feature representation for each base class ($i.e.$, prototype) while NCM is designed to enhance the transferability of such a learning ability, thereby reducing the testing bias. Although the idea of using the prototype as the class representation has been exploited in \cite{dong2018few, wang2019panet}, we propose to gradually update them during training for the first time, which leads to learning a more robust and discriminative representation. Moreover, POPNet adopts a stage-wise progressive human parsing framework, parsing human from the coarsest granularity to the finest granularity. Specifically, it incorporates the learned parent knowledge at the coarse granularity into the learning process at the fine granularity via a Knowledge Infusion Module (KIM), which enhances the discrimination of human part features for dealing with the small sizes issue.

The main contributions of this work are as follows. Firstly, we define a new and challenging task, $i.e.$, One-Shot Human Parsing, which brings new challenges and insights to the human parsing and one-shot learning community. Secondly, to address the problem, we propose a novel one-shot human parsing method named POPNet that is composed of a dual metric learning module, a dynamic human-part prototype generator, and a hierarchical progressive parsing structure that can address the testing bias and small sizes challenges. Finally, the experiments on the ATR-OS benchmark tailored for OSHP demonstrate our POPNet achieves superior performance than representative OS3 models and can serve as a strong baseline for the new problem.

\section{Related Work}

\subsection{Human Parsing} Human parsing aims at segmenting an image containing humans into semantic sub-parts including body parts and clothing classes. Recent success in deep CNN has made great progress in multiple areas \cite{ronneberger2015u, chen2017deeplab, zhan2020multi, zhang2020empowering, zhan2020multi, ma2020auto}, including human parsing \cite{li2017multiple, zhao2017self, luo2018macro}. Instead of tackling the human parsing task with a well-defined class set, we propose to solve a new and more challenging one named OSHP, which requires to parse human into an open set of classes with only one support example. Recent methods for human parsing improve parsing performance from utilizing the body structure priors and class relations \cite{xiao2018unified, gong2017look, zhu2018progressive, gong2018instance, li2020self, zhan2019exploring}. One direction is modeling the parsing task together with the keypoint detection task \cite{xia2017joint, nie2018human, huang2017coarse, fang2018weakly, dong2014towards, zhang2020correlating}. For example, Liang $et~al.$ proposed mutual supervision for both tasks and dynamically incorporated image-level context \cite{liang2018look}. The other direction is leveraging the hierarchical body structure at different granularities \cite{gong2019graphonomy, he2020grapy, wang2019learning, wang2020hierarchical}. For example, He $et~al.$ devised a graph pyramid mutual learning method to enhance features learned from different datasets with heterogeneous annotations \cite{he2020grapy}. In this spirit, we also use a hierarchical structure in our POPNet to leverage the learned knowledge at the coarse granularity to aid the learning process at the fine granularity, thereby enhancing the feature representation and discrimination especially in the one-shot setting.

\subsection{One-Shot Semantic Segmentation} One-Shot Semantic Segmentation (OS3) \cite{shaban2017one} aims to segment the novel object from the query image by referring to a single support image and the support object mask. Following the one/few-shot learning \cite{koch2015siamese, finn2017model, snell2017prototypical, sung2018learning, chen2020new, liu2020crnet, tian2020prior}, a typical OS3 solution is to learn a good metric \cite{zhang2018sg, rakelly2018conditional, zhang2019canet, zhang2019pyramid, hu2019attention, tian2020differentiable}. Zhang $et~al.$ extracted the target class centroid and calculated the cosine similarity scores as guidance to enhance the query image features and provided a strong metric \cite{zhang2018sg}. Recently, the metric was further improved by decomposing the class representations by part-aware prototypes in \cite{liu2020part}. Besides, comparing to the one-shot one-way setting, one-shot k-way semantic segmentation has also been studied by segmenting multiple classes at the same time \cite{dong2018few, wang2019panet, siam2019amp}. In contrast to the typical OS3 tasks where only intact objects of novel classes are presented and needed to be segmented, OSHP requires parsing human into small parts of both base classes and novel classes, which is similar to the challenging generalized few-shot learning (GFSL) setting tailored for practical usage scenarios \cite{gidaris2018dynamic, ren2019incremental, shi2019relational, ye2019learning}. To the best of our knowledge, GFSL for dense prediction tasks remains unexplored. In this paper, we make the first attempt by proposing a novel POPNet that employs a dual-metric learning strategy to enhance the transferability of the learning ability for recognizing human parts of base classes to novel classes.

\begin{figure*}[t]
  \centering
  \includegraphics[width=\linewidth]{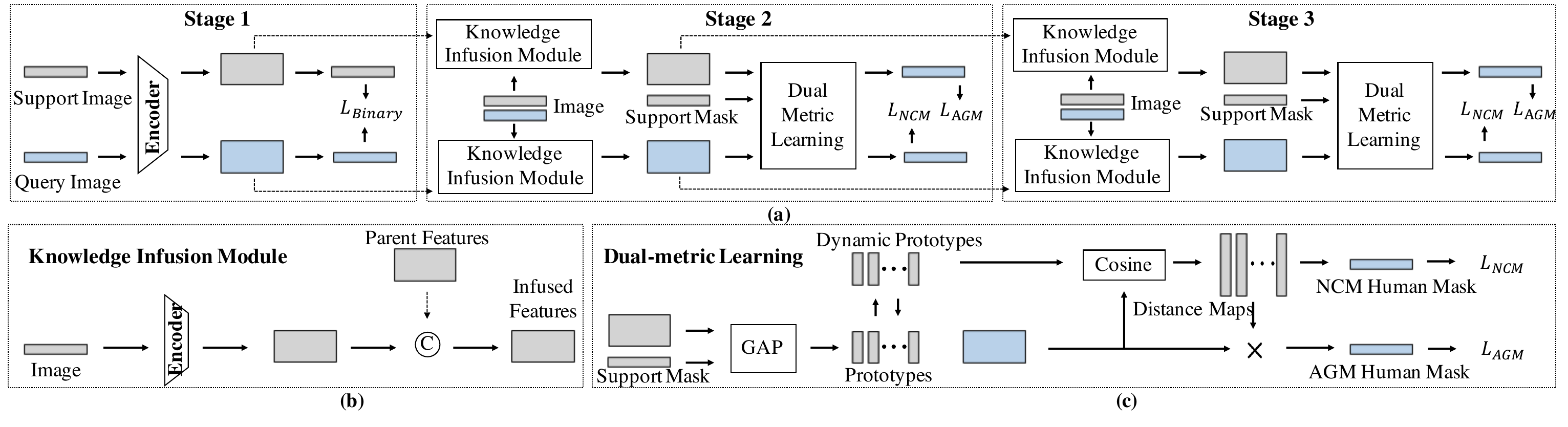}
  \caption{(a) Overview of the proposed three-stage POPNet. Each stage contains one encoder that embeds images into different semantic granularity level features. Stage 1, Stage 2, and Stage 3 generate foreground-background priors and masks, the main body area priors and masks, and the final fine-grind parsing masks respectively. (b) The structure of KIM. $C$ denotes feature concatenation. (c) The structure of Dual-Metric Learning. GAP represents class-wise global average pooling.}
  \label{fig:main}
\end{figure*}

\section{Problem Definition} \label{subsec:problemDef}
In this paper, we propose a new task named one-shot human parsing that requires to parse human in a query image according to the classes defined in a given support image with dense annotations. The training set is composed of many human images with dense annotations whose classes are partly overlapped with those in the support image. 

Using the meta-learning language \cite{shaban2017one, vinyals2016matching, zhang2018sg}, in the meta-training phase, a training set $\mathcal{D}_{train}$ along with a set of classes $C_{base}$ is given. In the meta-testing phase, the images in a test set $\mathcal{D}_{test}$ is segmented into multiple classes $C_{human} = C_{base} \cup C_{novel}$, where $C_{novel}$ is a flexible open set of classes that have never been seen during training and can be added or removed on the fly. Specifically, both sets in $\mathcal{D}$ consist of a support set and a query set. For meta-training, given $D_{train}$ of length $N$ annotated in $C_{base}$, the support set is denoted as $\mathcal{S}_{train} = \{(I^s, Y^s_{C_s}) | s\in \left[ {1,N_{train}} \right], C_s\subseteq C_{base}\}$, where $N_{train} < N$ is the number of training pairs, $Y^s_{C_s}$ is the ground truth support mask annotated in $|C_s|$ human parts defined in $C_s$. Similarly, the query set is denoted as $\mathcal{Q}_{train} = \{(I^q, Y^q_{C_s}) | q\in \left[ {N_{train}+1, N} \right], C_s\subseteq C_{base}\}$. Note that the query image is required to be parsed into the classes defined in $C_s$. For meta-testing, the support set $\mathcal{S}_{test}$ is similar to $\mathcal{S}_{train}$ except that the support masks are annotated in $C_{s} \subseteq C_{human}$. As for $Q_{test}$, only query images are provided, and the goal is to predict the query mask. During the meta-training phase, training pairs $\big((I^s, Y^s_{C_s}), (I^q, Y^q_{C_s})\big)$ are sampled from $\mathcal{S}_{train}$ and $\mathcal{Q}_{train}$ in each episode. The meta-learner aims to learn a mapping $\mathcal{F}$ subjected to $\mathcal{F} \big((I^s, Y^s_{C_s}), I^q\big) \to Y^q_{C_s}$ for any training pair. In the meta-testing phase, the meta-learner quickly adapts to any new pairs $\big((I^{s'}, Y^{s'}_{C_{s'}}), I^{q'}\big)$ sampled from $\mathcal{S}_{test}$ and $\mathcal{Q}_{test}$, $i.e.$, $\mathcal{F} (I^{s'}, Y^{s'}_{C_{s'}}) \to Y^{q'}_{C_{s'}}$.

It's noteworthy that the classes in $C_{test}$ are not necessarily connected regarding the human body structure. For example, the model can be trained with some base classes like arms and legs and evaluated with some novel classes like shoes and hat. However, we argue that given some base classes that have strong correlations with the novel ones, for example, legs and pants, it is easy to infer the novel classes like shoes in the meta-testing phase. To better evaluate the transferability of the model's learning ability to novel classes, we split the $C_{train}$ and $C_{test}$ to be cluster-disjoint manner which means that all the subclasses belonging to the same parent class should be in the same set. For example, $C_{train}$ may contain hair and face (in the `head' parent class), while $C_{test}$ contains legs and shoes (in the `leg' parent class). Obviously, this setting is more challenging.

\section{Progressive One-shot Parsing Network}

To address the two key challenges in OSHP, we devise a POPNet. It has three stages of different granularity levels (Figure~\ref{fig:main} (a)). The first stage generates the foreground body masks, the second stage generates the coarse main body area masks, and the third stage generates the final fine-grind parsing masks. The learned semantic knowledge of each stage is inherited by the next stage via a Knowledge Infusion Module (Figure~\ref{fig:main} (b)) to enhance the discrimination of human part features and deal with the small sizes issue. In the second and third stage, a Dual-metric Learning (DML)-based meta-learning method ((Figure~\ref{fig:main} (c))) is proposed to generate robust dynamic class prototypes that can generalize to the novel classes thereby reducing the testing bias.

\subsection{Progressive Human Parsing} \label{subsec:progressive}
Instead of being intact objects, human parsing classes are non-holistic small human parts, which makes it non-trivial to directly adopt the OS3 metric learning methods on the OSHP task. Inspired by \cite{he2020grapy}, the human body is highly structural and the semantics at human coarse granularity can help network eliminate distractions and focus on the target classes at the fine granularity. To this end, we decompose our POPNet into three stages from the coarse granularity to the fine granularity. By infusing the learned knowledge from the coarse stages into the fine stage via a knowledge infusion module (detailed in Section~\ref{subsec:kim}), the network boosts the pixel-wise feature representations with rich parents semantics to discriminate the small-sized human parts.

Specifically, in one episode, we are provided with the query image $I^q \in \mathbb{R}^{H \times W \times 3}$, support image $I^s \in \mathbb{R}^{H \times W \times 3}$ and support mask $Y^s_{C_s} \in \mathbb{R}^{H \times W \times |C_s|}$ that annotated in class set $C_s$. The network's expected outcome is the predicted $Y^q_{C_s} \in \mathbb{R}^{H \times W \times |C_s|}$ which assigns the classes in the support mask to the query image pixels. In the first stage, we devise a binary human parser that can segment the human foreground out of the background. It is trained via supervised learning by leveraging additional binary masks $Y^q_{C_{fg}}$ and $Y^s_{C_{fg}}$, $i.e.$, $|C_{fg}| = 2$, derived from $Y^q_{C}$ and $Y^s_{C}$ by replacing all the foreground classes with a single foreground label. Noting that here we adopt the conventional supervised foreground segmentation setting instead of the one-shot one-way foreground segmentation setting since a well-trained human foreground parser can include most of the possible human-related classes and cause no harm to the potential novel classes semantically. Besides, there is no large-scale one-shot segmentation dataset that contains the human class while having a small domain gap with the existing human parsing datasets. We leave it as the future work to explore the one-shot one-way setting in the first stage.

In the second stage, we follow the OSHP settings and devise a one-shot meta learner on the main body areas that the parsed foreground classes in this stage are at the coarse granularity, $i.e.$, head, body, arms, legs, and the background. Assuming $C_{\hat{s}}$ is the set of the main body areas, we can get the supervision $Y^s_{C_{\hat{s}}}$ by aggregating $Y^s_{C_s}$, $i.e.$, replacing the class labels belonging to the same parent class with the parent class label. Hence, $|C_{\hat{s}}| = 4$ and $|C_{\hat{s}}| = 5$ during the meta-training and meta-testing respectively, since one coarse class serves as the novel class during training according to 
Section.~\ref{subsec:problemDef}. Accordingly, the model learns the body semantics via the meta-learning in the second stage. In the third stage, we devise the one-shot human parsing meta learner that can predict the fine-granularity human classes $Y^q_{C_s}$.

\subsection{Knowledge Infusion Module} \label{subsec:kim}
To fully exploit context information from the the previous stages and enhance the representative ability of the features in the current stage, we propose to infuse the learned parent knowledge into learning process when inferring its descendants. Specifically, in 
each stage, the input image (query image or support image) is fed into a shared encoder network to get the embedded features $g^{S_i}, i=1,2,3$. In the second stage, we exploit $g^{S_1}$ by concatenating it with the image features learned in the second stage $g^{S_2}$ to get the enhanced features $h^{S_{2}}$ via a knowledge infusion module, $i.e.$, $h^{S_{2}} = \zeta_2 \left([g^{S_{1}}; g^{S_{2}}]\right),$ where $[;]$ denotes the concatenation operator, $\zeta_2$ represents the mapping function learned by two consecutive conv layers. Likewise, in the third stage,  we exploit $h^{S_2}$ by concatenating it with the image features learned in the third stage $g^{S_3}$ to get the enhanced features $h^{S_{3}}$, $i.e.$, $h^{S_{3}} = \zeta_3 \left([h^{S_{2}}; g^{S_{3}}]\right)$. All encoded features and infused features are in $\mathbb{R}^{H \times W \times K}$, where $K$ denotes feature channels. 

We implement the encoder in each stage using the Deeplab v3+ model \cite{chen2018encoder} with an Xception backbone \cite{chollet2017xception}. We use the features before the classification layer as $g^{S_i}$, since they contain semantic-related information. In this way, the learned hierarchical body structure knowledge is infused into the next stage progressively to help discriminate the fine-granularity classes via dual-metric learning (detailed in Section~\ref{subsec:DML}). We train the three stages sequentially and fix the model parameters in the previous stage when training the current stage.

\subsection{Dual-Metric Learning}
\label{subsec:DML}
Take the third stage as an example, given the support mask $Y^s_{C_s}$, infused features $h^s$ and $h^q$,
it is desired to generate the query mask $Y^q_{C_s}$. In the OS3 methods, inferring the query mask is accomplished by using convolution layers \cite{hu2019attention, gairola2020simpropnet, zhang2019canet} or graph reasoning \cite{zhang2019pyramid} to explore pixel relationships. Recently, \cite{tian2020prior} propose to solve support-query inconsistency by enriching the features through a pyramid-alike structure. The mentioned approaches can be summarized as post feature enhancement approaches that learn the implicit query-support correlations after the encoder. However, the learned query-support correlations in the enriched features are likely to be overfitting on the base class, thereby reducing the transferability on the generalized OSHP setting. In this case, we choose the simple yet effective design that computes the cosine similarity scores \cite{zhang2018sg, liu2020part} between the features and class prototypes, which shows a better transferability.

\subsubsection{Dynamic Prototype Generation} 
\label{subsec:dpg}
Different to the prior prototype methods, we propose to generate more robust dynamic prototypes. First, we calculate the class prototype $p_{c}$ for class $c \in C_s \setminus c_{bg} $ \cite{zhang2018sg} as:
\begin{equation}
p_{c} = \frac{1}{{\left| {\Lambda _{c}} \right|}}\sum\limits_{\left( {x,y} \right) \in \Lambda _{c}} {{h^{s}}\left( {x,y} \right)},
\label{eq:pix2proto}
\end{equation}
where $(x, y)$ denote pixel index, $\Lambda_{c}$ is the support mask of class $c$, $\left| {\Lambda _{c}} \right|$ is the number of pixels in the mask. Note that in the prior methods \cite{ dong2018few} that a `background prototype' is learned to represent non-foreground regions. However, in the OSHP setting, the background pixels in the training data include both background and the novel classes in $C_{novel}$. Therefore, we do not calculate the background prototype to prevent pushing the novel classes towards background class. Instead, we predict the background by excluding all the foreground classes in the following sessions.

Instead of using a static $p_c$ in the following networks, we generate a dynamic prototype ${p}^d_c$ to improve the robustness of base class representation. Specifically, it is calculated by gradually smoothing the previous prototype estimate ${p}^d_c$ and the current estimate $p_c$ in each episode, $i.e.$,
\begin{equation}
{p}^d_c = \alpha \times {p}^d_c + (1-\alpha) \times p_c,
\label{eq:dynamicPrototype}
\end{equation}
where $\alpha$ is the smoothing parameter. Since the novel classes are not seen in training, we use static prototypes for the novel classes during testing. For simplicity, we denote both prototypes as $\hat{p}$ in the following sessions. Next, the distance map $m_c$ between the query features ${h_{q}}$ and the class prototype is calculated by cosine similarity as follows: $m_{c} = <{h^{q}}, \hat{p}_{c}>$. Prior methods mainly utilize distance maps in two ways. Parametric approach \cite{zhang2018sg} uses distance maps as the attention by element-wise multiplying the distance map to the query image feature maps for further prediction. Non-parametric approach \cite{wang2019panet} directly makes predictions basing on the distance maps. We find that the first approach can learn a better metric on the base classes while cannot generalize well on the novel classes. In contrast, the second approach has a strong transferability due to the effective and simple distance metric, but it struggles to discriminate the human classes that are semantically similar. To this end, we propose a novel weight shifting strategy for DML such that it disentangles metric's representation ability and model's generalization ability. In the early training phase, DML learns the metric for better representation using AGM. In the late phase, DML shifts focus to improve the transferability of this learning ability on novel classes and addresses testing bias issue using NCM. 

\subsubsection{Attention Guidance Module} \label{subsubsec:AGM} 
In the early training phase, our meta learner aims to fully exploit the supervisory signals from base classes and learn a good feature representation. To this end, we use the distance maps $m_{c}$ as the class-wise attention to enhance the query features in a residual learning way, $i.e.$, $r_c = m_c \times h^{q} + h^{q}$. Then, we generate probability map for each class by feeding $r_c$ to a convolutional layer $\varphi$ and a softmax layer, $i.e.$, 
\begin{small}
\begin{equation}
    \begin{aligned}
     & Y_{c}^{q;AGM} = \frac{exp(\varphi(r_c))}{\sum_{c\in C_s \setminus c_{bg}}exp(\varphi(r_c)) + r_{bg}} \\
     & r_{bg} = (1/(|C_s| - 1)) \times \sum_{c\in C_s\setminus c_{bg}}\omega (r_c).
    \end{aligned}
\label{eq:agm2}
\end{equation}
\end{small}
Note that we infer the probability map for the background class by aggregating all the foreground features after a convolutional layer $\omega$, which can automatically attend to the non-foreground regions by learning negative weights.

\subsubsection{Nearest Centroid Module} \label{subsubsec:NCM} 
In the late training phase, our meta learner aims to increase the transferability of the learning ability from base classes to novel classes. To this end, we propose the non-parametric Nearest Centroid Module that infers the probability map directly from the similarity between features and class prototypes. Specifically, we use a softmax layer directly on the distance maps $m_c$ and $m_{bg}$ to get the final prediction. Likewise, we get $m_{bg}$ by explicitly averaging all the reverse foreground distance maps, $i.e.$,
\begin{small}
\begin{equation}
    \begin{aligned}
    & Y_{c}^{q;NCM} = \frac{exp(m_c)}{\sum_{c\in C_s\setminus c_{bg}}exp(m_c) + m_{bg}} \\
     &m_{bg} = (1/(|C_s| - 1)) \times \sum_{c \in C_s\setminus c_{bg}} (1 - m_c).
    \end{aligned}
\label{eq:ncm}
\end{equation}
\end{small}
\subsubsection{Weight Shifting Strategy} During training, we control the meta-learner's focus by assigning dynamic loss weights for both modules, $i.e.$,

\begin{equation}
\begin{small}
    \begin{aligned}
     L^{dml} &= -\sum_{(x,y)=1}^{N_p}\sum_{c=1}^{|C_s|} \Big[\beta \times y_c { log}\left( \hat{y}_c^{q;{ AGM}}\left( x, y\right) \right) \\
    +&
    \left( 1 - \beta \right) \times y_c{ log} \left (\hat{y}_c^{q;{ NCM}} \left( x, y \right) \right) \Big],
    \end{aligned}
\label{eq:weight}
\end{small}
\end{equation}
where $\hat{y}_c^{q; AGM}(x,y)$ and $\hat{y}_c^{q; NCM}(x,y)$ are one pixel in $\hat{Y}_c^{q; AGM}$ and one pixel in $\hat{Y}_c^{q; NCM}$ indexed by $(x,y)$ 
respectively, $N_p$ is the numbers of pixels in $\hat{Y}_c^{q; AGM}$ and $\hat{Y}_c^{q; NCM}$,
$y_c$ is a binary indicator function outputting $1$ when class $c$ is the target class, and $\beta$ is a balancing hyperparameter. During training, we linearly decrease $\beta$ by $\beta = 1 - {epoch / max\_epoch}$ thus gradually shifting network's focus from AGM to NCM.

\section{Dataset and Metric}
\subsubsection{Dataset: ATR-OS} In this session, we illustrate how to tailor the existing large-scale ATR dataset \cite{liang2015, liang2015human} into a new ATR-OS dataset for the OSHP setting. We choose the ATR dataset instead of the MHP dataset \cite{li2017multiple, zhao2018understanding} for the following reasons. First, ATR dataset a large-scale benchmark including 18000 images annotated with 17 foreground classes. The abundant labeled data allow the network to learn rich feature representations. Second, ATR's images are mostly fashion photographs including models and a variety of fashion items, which are closely related to OSHP's applications such as fashion clothing parsing \cite{yamaguchi2012parsing}. Third, comparing to the other datasets, models' poses, sizes, and positions in the ATR dataset have less diversity. Hence, it is a good start for the newly proposed challenging OSHP task. We leave the research on OSHP in complex scenes as future work.

We split the ATR samples into support sets and query sets according to the one-shot learning setting for training and testing respectively. We form $\mathcal{Q}_{train}$ by including the first 8000 images of the ATR training set and form $\mathcal{S}_{train}$ with the remaining images. We form the 500-image $\mathcal{Q}_{test}$ and 500-image $\mathcal{S}_{test}$ from the original test set in a similar way. In each training episode, we randomly select one query-support pair from $\mathcal{Q}_{train}$ and $\mathcal{S}_{train}$, while in each testing episode, the network is evaluated by mapping each sample from $\mathcal{Q}_{test}$ to 10 support samples from $\mathcal{S}_{test}$ and forms a 5000 testing pairs in total. For a fair comparison, the 10 support samples are fixed. The selection for $\mathcal{S}_{test}$ images is illustrated in supplementary materials.

To ease the difficulty for training OSHP on the ATR dataset, we merge the symmetric classes and rare classes in ATR, $e.g.$ `left leg' and `right leg' are merged as `legs' and `sunglasses' is merged into the background. Before training, the remaining 12 classes including `background' denoted as $C_{human}$ are sampled into $C_{base}$ and $C_{novel}$. To limit the networks to only learn from the classes in $C_{base}$ during training, the regions of $C_{novel}$ are merged into `background', thereby only classes in $C_{base}$ are seen in $D_{train}$. During testing, all classes indicated by the support masks are evaluated, including classes from both $C_{base}$ and $C_{novel}$. Note that it is unreasonable in a query-support pair that some classes required to be parsed in the query image are not annotated in the support mask, so we merge these classes into `background' as well. Besides, due to the reason illustrated in Section~\ref{subsec:problemDef}, $C_{novel}$ is chosen from the two sets representing two main body areas, respectively, $i.e.$, the leg area: $C_{Fold\ 1}$ = [pants, legs, shoes] and the head area: $C_{Fold\ 2}$ = [hair, head, hat]. 

\subsubsection{Metrics} 
We use Mean Intersection over Union (MIoU) as the main metric for evaluating the parsing performance on the novel classes $C_{novel}$ and all the human classes $C_{human}$. We also compute average overall accuracy to evaluate the overall human parsing performance. For the one-way setting as described in Section~\ref{subsec:expMain}, we also compute the average Binary-IoU \cite{wang2019panet}. To avoid confusion, we refer to the main evaluation setting as k-way OSHP that parses k human parts at the same time while we refer to parsing only one class in each episode as one-way OSHP.

\begin{table}[]
\centering
\resizebox{0.45\textwidth}{!}{%
\fontsize{15pt}{15pt}\selectfont
\begin{tabular}{@{}cccccccc@{}}
\toprule
\multirow{2}{*}{Method} & \multicolumn{3}{c}{Novel Class MIoU} & \multicolumn{3}{c}{Human MIoU} & \multirow{2}{*}{Overall Acc} \\ \cmidrule(lr){2-7}
       & Fold 1 & Fold 2 & Mean & Fold 1 & Fold 2 & Mean &      \\ \midrule
AMP    & 8.5    & 8.1    & 8.3  & 16.3   & 15.4   & 15.9 & 67.6 \\
SG-One & 0.0    & 0.1    & 0.1  & 42.7   & 46.0   & 44.4 & 91.6 \\
PANet  & 12.6   & 13.3   & 13.0 & 19.4   & 17.1   & 18.3 & 78.8 \\ \midrule
POPNet & \textbf{24.1}   & \textbf{19.4 }  & \textbf{21.8} & \textbf{60.6}   & \textbf{60.4}   & \textbf{60.5} & \textbf{94.1} \\ \bottomrule
\end{tabular}%
}
\caption{Comparison on k-way OSHP with the baselines. Human MIoU refers to the MIoU on $C_{human}$.}
\label{tab:k-way}
\end{table}

\begin{table}[]
\centering
\resizebox{0.45\textwidth}{!}{%
\fontsize{15pt}{15pt}\selectfont
\begin{tabular}{@{}cccccccc@{}}
\toprule
\multirow{2}{*}{Method} & \multicolumn{3}{c}{Novel Class MIoU} & \multicolumn{3}{c}{Human MIoU} & \multirow{2}{*}{Bi-IoU} \\ \cmidrule(lr){2-7}
       & Fold 1 & Fold 2 & Mean & Fold 1 & Fold 2 & Mean &      \\ \midrule
Fine-tune & 0.3    & 0.2    & 0.3  & 14.8   & 15.0   & 14.9 & 49.1 \\
AMP    & 8.4    & 9.4    & 8.8  & 15.0   & 15.0   & 15.0 & 50.7 \\
SG-One & 4.0    & 0.7    & 2.4  & 39.0   & 40.5   & 39.8 & 66.0 \\
PANet  & 5.1   & 3.2   & 4.2 & 14.0   & 13.9   & 14.0 & 49.5 \\ \midrule
POPNet & \textbf{28.3}   & \textbf{28.4}   & \textbf{27.7} & \textbf{51.1}   & \textbf{54.6}   & \textbf{52.8} & \textbf{71.4} \\ \bottomrule
\end{tabular}%
}
\caption{Comparison on one-way OSHP with the baselines.}
\label{tab:1-way}
\end{table}

\section{Experiments}
\subsection{Baselines}
\label{subsec:baselines}
\textbf{Fine-Tuning}: as suggested in \cite{caelles2017one}, we first pre-train the model on $\mathcal{D}_{train}$ then fine-tune on the $\mathcal{S}_{test}$ for a few iterations. Specifically, we use the same backbone as POPNet and only fine-tune the last two convolution layers and the classification layer. \textbf{SG-One}: we follow the settings of SG-One \cite{zhang2018sg} and learn similarity guidance from the support image features and support mask. We use the same backbone as POPNet for better performance. To support k-way OSHP, we follow a similar prediction procedure as defined by Eq.~\eqref{eq:agm2} in our AGM except that it does not use residual learning. 
\textbf{PANet}: we use PANet as another baseline with non-parameter metric learning and prototype alignment loss. In the k-way OSHP, we pair each query image with k support images that each contains a unique class ($i.e.$, with a binary support mask) in the support set as is described in \cite{wang2019panet}. \textbf{AMP}: we use masked proxies with multi-resolution weight imprinting technology and carefully tune a suitable learning rate as described in \cite{siam2019amp}.

\begin{figure}[t]
  \centering
  \includegraphics[width=\linewidth]{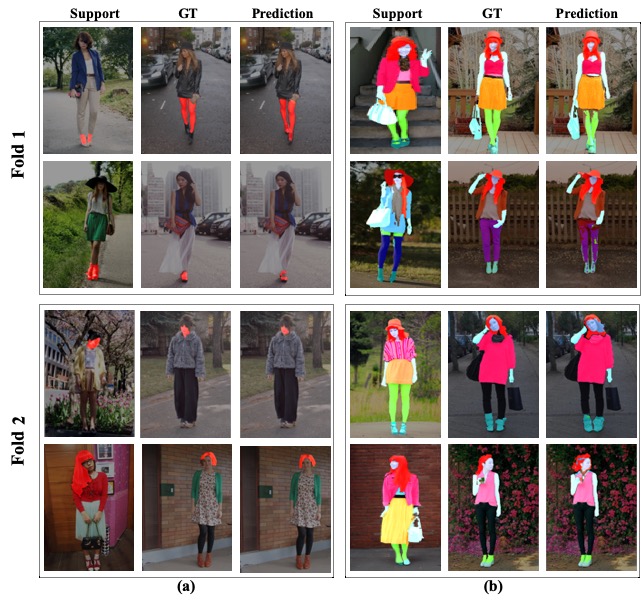}
  \caption{Visual results on ATR-OS. (a) One-way OSHP. (b) K-way OSHP.}
  \label{fig:visual}
\end{figure}
\subsection{Implementation Details}
In this paper, we conduct the experiments on a single NVIDIA Tesla V100 GPU. The backbone network is pre-trained on the COCO dataset \cite{lin2014microsoft}. The images are resized to $576 \times 576$ in one-way OSHP tasks and resized to $512 \times 512$ in k-way tasks due to memory limit, which leads to computations of 131.3 GMacs and 100.4 GMacs respectively. Training images are augmented by a random scale from 0.5 to 2, random crop, and random flip. We train the model using the SGD optimizer for 30 epochs with the poly learning rate policy. The initial learning rate is set to 0.001 with batch size 2. When generating dynamic prototypes, $\alpha$ in Eq.~\eqref{eq:dynamicPrototype} is set to 0.001 by grid search. However, static prototypes are utilized when calculating distance maps in the first 15 epochs before we aggregating enough prototypes to reduce the variance and get stable dynamic prototypes.


\subsection{Results and Analysis}
\label{subsec:expMain}
\subsubsection{Comparison With Baselines} \textbf{K-Way OSHP}: we compare our model with the customized OS3 baseline models described in Table~\ref{tab:k-way}. We first report the results on the overall human classes, POPNet significantly improves the Human MIoU and Overall Accuracy to 60.5\% and 94.1\% respectively, which demonstrates the three-stage progressive parsing can develop pixel-wise feature representations with rich semantic that address the small-sized human part issue. When evaluating the novel classes, our method has outperformed the baseline methods by 8.8\%. The significant margin demonstrates that the class centroid learned by dynamic prototypes in DML can successfully be generalized to novel classes and reduce the testing bias effect. 

\textbf{One-Way OSHP}: in addition to k-way OSHP, we also report results on one-way OSHP that the support mask only contains one class in one episode. In this setting, we evaluate each class 500 times with different $(s, q)$ pairs randomly sampled from $\mathcal{S}$ and $\mathcal{Q}$. As is seen from Table~\ref{tab:1-way}, our model achieves 27.7\%, 52.8\%, and 71.4\% in mean novel class MIoU, mean human MIoU, and mean Bi-IoU and outperforms the best baseline method by a margin of 18.9\%, 13.0\%, and 5.4\% respectively. POPNet's superiority in solving small human parts and novel classes is again confirmed by the large margins on the one-way OSHP. Note that when comparing one-way OSHP scores to k-way OSHP, the novel class MIoU is higher while the overall human MIoU is lower. The reason is that the model would be less confident in the novel classes when the base classes are involved in the testing at the same time. However, multiple classes' prototypes from the k-way supervisory signals would help our model learn the underlying human part semantic relations and improve the overall human parsing performance.


\begin{figure}[t]
  \centering
  \includegraphics[width=\linewidth]{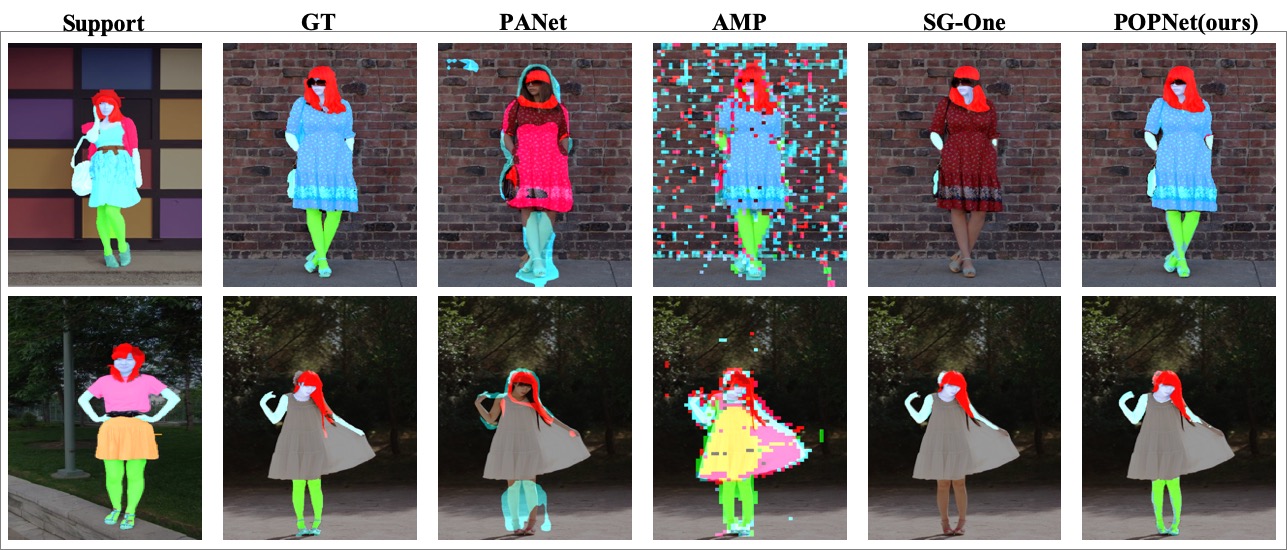}
  \caption{Visual comparison with baselines on ATR-OS.}
  \label{fig:visual2}
\end{figure}

\subsubsection{Visual Inspection} To better understand the OSHP task and POPNet, we show the qualitative results on both OSHP settings in Figure~\ref{fig:visual}. In one-way OSHP, when given image-mask support pair on one novel class, $e.g.$ `shoes' in Fold 1, POPNet can segment the part mask of the novel class from the query image accurately although the appearance gap is huge. In k-way OSHP, when the full human mask is given, POPNet can efficiently parse human into multiple human parts including both base and novel classes. The qualitative results show that our method can flexibly generate satisfying parsing masks with classes defined by the support example.

We further compare the visual results with the baseline methods on ATR-OS Fold 1 in Figure~\ref{fig:visual2}. As is seen, the baseline methods struggle in recognizing the small-sized human parts and are only able to separate the holistic human foreground from the background. Although SG-One \cite{zhang2018sg} can segment the pixels of base classes, $e.g.$ `hair', it tends to overfit these base classes and cannot find the novel classes. In contrast, POPNet can discriminate and parse the non-holistic small classes by reducing the testing bias effect via DML and tackling the small-sized human parts with the progressive three-stage structure.

\subsubsection{Ablation Study} We investigate the effectiveness of key POPNet components in this session. Firstly, as in Table~\ref{tab:ablation}, 
when applying either AGM or NCM on the ATR-OS dataset ($\beta=0$ or $\beta=1$ in Eq~\eqref{eq:weight}), the two models only achieve less than 5\% novel class MIoU and around 40\% human MIoU. The scores suggest that the models can only recognize the holistic human contour and cannot segment the novel classes due to the small sizes and testing bias challenges. Next, we evaluate the weight shifting effect by comparing DML with and without weight shifting. DML without weight shifting ($\beta=0.5$ in Eq~\eqref{eq:weight}) utilizes both metric learning modules and achieves 15.1\% novel class MIoU and 47.9\% human MIoU, much better than any single module. By using the weight-shifting strategy, our model can significantly reduce the testing bias, improving the novel class MIoU by a margin of 3.5\%. It validates that shifting the network's focus on NCM during the late stage of training can considerably improve the network's transferability to novel classes. We then apply progressive human parsing on the existing structure, the human MIoU is noticeably improved by 5.4\%, and novel class MIoU is improved to 20\%, which demonstrates that employing hierarchical human structure is beneficial for tackling the small-sized human parts. Finally, POPNet can achieve 24.1\% novel class MIoU and 60.6\% human MIoU through dynamic prototype generation. In addition to the table content, the base class MIoU from $C_{human} \setminus C_{novel}$ reaches 72.8\%, which is close to the fully supervised human parsing methods. It is indicated that increasing the robustness of the base class representation is also helpful for transferring the knowledge to the novel classes and gaining a remarkable margin on the novel class MIoU. 

\begin{table}
\resizebox{\linewidth}{!}{
\centering
\small
\begin{tabular}{@{}ccc@{}}
\toprule
Methods   & Novel Class MIoU & Human MIoU \\ \midrule
AGM  & 1.0  & 40.2 \\
NCM  & 3.5  & 40.4  \\
DML & 15.1 & 47.9    \\
DML + WS & 18.6 & 47.1   \\                 
DML + WS + KIM & 20.0                     &  52.5              \\
DML + WS + KIM + DP  & 24.1                      &  60.6                 \\ \bottomrule
\end{tabular}%
}
\caption{Ablation study on ATR-OS Fold 1. WS is short for the weight shifting strategy, KIM is short for the three-stage progressive human parsing with knowledge infusion module, and DP is short for the dynamic prototype generation.}
\label{tab:ablation}
\end{table}

\subsubsection{Limitation} We find that there is still a performance gap between the novel classes and the base classes. Such gap mainly comes from the confusion among the novel classes, $e.g.$ shoes, and legs in Figure~\ref{fig:visual} (a). To address this issue, modeling the inter-class relations using graph neural network and reasoning on the graph may enhance the feature discrimination further, which we leave as the future work. 

\section{Conclusion}
We introduce a new challenging but promising problem, $i.e.$, one-shot human parsing, which requires parsing human into an open set of classes defined by a single reference image. Moreover, we make the first attempt to build a strong baseline, $i.e.$, Progressive One-shot Parsing Network (POPNet). POPNet adopts a dual-metric learning strategy based on dynamic prototype generation, which demonstrates its effectiveness for transferring the learning ability from base seen classes to novel unseen classes. Moreover, the progressive parsing framework effectively leverages human part knowledge learned at the coarse granularity to aid the feature learning at the fine granularity, thereby enhancing the feature representations and discrimination for small human parts. We also tailor the popular ATR dataset to the one-shot human parsing settings and compare POPNet with other representative one-shot semantic segmentation models. Experiment results confirm that POPNet outperforms other models in terms of both generalization ability on the novel classes and the overall parsing ability on the entire human body. The future work may include 1) constructing new benchmarks for this task by paying more attention to appearance diversity, $e.g.$, pose and occlusion; and 2) modeling inter-class correlations to enhance the feature representations. 

\section{Acknowledgements}
This work was supported by the Australian Research Council Projects FL-170100117, DP-180103424, IH-180100002.

\bibstyle{aaai21}
{
\begin{small}
\bibliography{reference}
\end{small}}

\end{document}